\documentclass[runningheads]{llncs}
\usepackage{graphicx}
\usepackage{amsmath}
\usepackage{amsfonts}
\usepackage{orcidlink}

\usepackage{hhline}
\usepackage{multirow}
\usepackage{makecell}
\usepackage{booktabs}
\usepackage{subcaption}

\begin{document}
\title{Ensembled Cold-Diffusion Restorations for Unsupervised Anomaly Detection}
%
%
\author{Sergio Naval Marimont\inst{1}\orcidlink{0000-0002-7075-5586} 
\and Vasilis Siomos\inst{1}\orcidlink{0009-0003-0985-2672}
\and Matthew Baugh\inst{2}\orcidlink{0000-0001-6252-7658}
\and Christos Tzelepis\inst{1}\orcidlink{0000-0002-2036-9089}
\and Bernhard Kainz\inst{2,3}\orcidlink{0000-0002-7813-5023} 
\and Giacomo Tarroni \inst{1,2}\orcidlink{0000-0002-0341-6138}
}
\authorrunning{S. Naval Marimont et al.}
%
\institute{City, University of London, UK \and Imperial College London, UK  \and Friedrich--Alexander University Erlangen--N\"urnberg, DE \\
\email{sergio.naval-marimont@city.ac.uk}}

\maketitle              
\begin{abstract}

Unsupervised Anomaly Detection (UAD) methods aim to identify anomalies in test samples comparing them with a normative distribution learned from a dataset known to be anomaly-free. Approaches based on generative models offer interpretability by generating anomaly-free versions of test images, but are typically unable to identify subtle anomalies. Alternatively, approaches using feature modelling or self-supervised methods, such as the ones relying on synthetically generated anomalies, do not provide out-of-the-box interpretability. In this work, we present a novel method that combines the strengths of both strategies: a generative \emph{cold-diffusion} pipeline (i.e., a diffusion-like pipeline which uses corruptions not based on noise) that is trained with the objective of turning synthetically-corrupted images back to their normal, original appearance. To support our pipeline we introduce a novel synthetic anomaly generation procedure, called DAG, and a novel anomaly score which ensembles restorations conditioned with different degrees of abnormality. Our method surpasses the prior state-of-the art for unsupervised anomaly detection in three different Brain MRI datasets.

\end{abstract}

\section{Introduction}
Unsupervised Anomaly Detection (UAD) in medical images involves the detection and/or localization of anomalies without requiring annotations, leveraging only a dataset known to be anomaly-free that \textit{describes} the so-called \textit{normative} distribution. At inference time, UAD methods try to assess whether a given test image belongs to the learnt normative distribution or not.
The interest for UAD stems from the fact that differently from common supervised approaches UAD methods 1) do not require annotations, which are expensive and challenging to obtain~\cite{Litjens2017}, and 2) are not limited to the labelled classes seen during training.

Multiple strategies have been proposed in the literature. Commonly, UAD methods start by defining a model of the observed normative distribution, either using generative models or self-supervised learning. Methods relying on generative models often \textit{query} the model to produce an image consistent with a given test image but without anomalies present, and identify as anomalous the areas where the generated image is different from the input test one \cite{Baur2020}. While generative models produce a useful healthy counterfactual, their UAD performance is usually limited by blurry reconstructions or lack of coverage of the healthy distribution \cite{UPD}. Instead, self-supervised strategies typically introduce synthetic anomalies into healthy samples and subsequently train a segmentation network to directly identify the introduced anomalies~\cite{FPI}. Under certain conditions, models are expected to generalize to unseen, naturally occurring anomalies. While very successful in recent editions of MICCAI's Medical Out-of-Distribution Challenge (MOOD)\cite{MOOD}, synthetic anomaly models lack in interpretability. 

Very recently, a novel strategy called Diffusion-Inspired Synthetic Restoration (DISYRE)~\cite{DISYRE} combining generative and synthetic anomaly methods have been presented. In this paper, we examine and address the limitations of previously proposed approaches, and implement a hybrid method inspired by diffusion models to turn synthetically corrupted images into healthy ones. Differently from previous approaches, we specifically introduce:
\begin{itemize}
    \item A novel synthetic anomaly generation process that increases the coverage of the anomaly distribution by disentangling attributes of the generated anomalies, and more importantly, makes the learnt model robust to the assumptions of severity which are required for inference.
    \item A novel method to identify anomalies which ensembles predictions created under different assumptions on severity.
\end{itemize}
We evaluated our novel hybrid method named DISYRE v2, in three Brain MRI datasets. DISYRE v2 substantially outperforms all previous methods, setting a new state-of-the-art in UAD. The codebase for DISYRE v2 is available online\footnote{https://github.com/snavalm/disyre.}.

\begin{figure*}[t!]
    \centering
    \includegraphics[width=1\columnwidth]{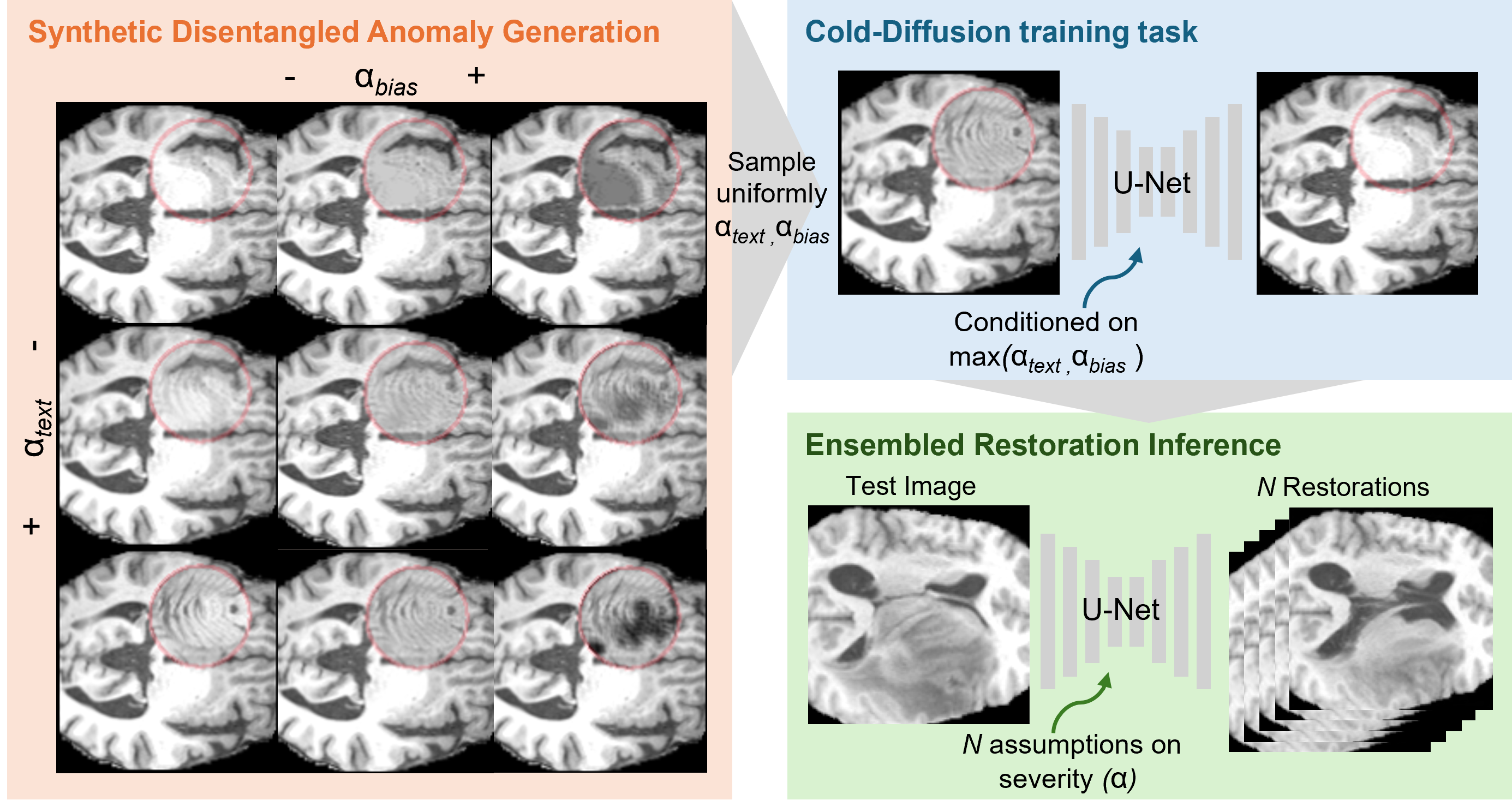}
    \caption{Overview of the main components of DISYRE v2, our proposed unsupervised anomaly detection pipeline.}
    \label{fig:anom_gen}
\end{figure*}

\section{Related Works}

\label{sections:related}
Synthetic anomaly UAD strategies consist in training segmentation networks to identify synthetically generated anomalies, expecting the model to generalize to unseen, naturally occurring abnormalities. 
First proposed in Foreign Patch Interpolation (FPI)~\cite{FPI}, authors realised that the key to generalization is the anomaly generation (AG) procedure, which needs to yield plausible anomalies with a wide coverage of the unknown anomalous distribution. In FPI anomalies are generated by interpolating a squared region of the sample with a patch extracted from a separate sample. A scalar $\alpha$  defines the amount of interpolation used in the corruption and can be used as a proxy for the anomaly severity. A segmentation network is then trained to localize the corruption. 
Subsequent works increased performance by seamlessly blending the anomalies with the surrounding area \cite{PII,MOOD}. More recent works \cite{nsa,manytasks} improved generalization imposing foreground conditions on the generated anomalies, proposing better training losses for the task and ensembling models trained on different synthetic tasks.

On a separate image modelling strategy, diffusion models have recently become the state-of-the-art generative models. Typically, Denoising Diffusion Probabilistic Models (DDPM)~\cite{DDPM} are trained to reverse a forward process by which images are gradually corrupted with Gaussian noise. By learning to \textit{undo} the noise corruption, diffusion models learn a score function, $\nabla_x \log p(x)$, which specifies how to modify image pixels to increase their joint probability $p(x)$. Image generation is achieved by iteratively denoising Gaussian noise. Cold-diffusion~\cite{cold_diffusion} follows the same concept, but replaces Gaussian noise with alternative corruptions (e.g.,  blurring, masking). Diffusion models have become recently popular in UAD: the most common approach is to directly train a diffusion model on anomaly-free images. At inference time, test images are corrupted with noise and subsequently denoised, aiming at obtaining an image restoration~\cite{Chen2020}, i.e. an image consistent with the test image but without the original anomalies \cite{anoddpm,pinaya_fast_2022}. Anomalies are then identified by comparing test images with their restorations. After corrupting test images with noise, restorations unavoidably shift from the original images, typically producing undesirable changes to low-frequency details, which makes these methods unable to identify subtle anomalies. 

Recently an approach named DISYRE~\cite{DISYRE} showed that, in principle, a score function can act directly as a pixel-wise anomaly score for UAD. Accordingly, a cold-diffusion pipeline with a gradual, synthetic anomaly corruption was proposed, so the learnt score function can generalize to unseen medical anomalies. Similarly to standard diffusion, models are conditioned with a time step $t$ which in DISYRE is a proxy for anomaly severity. Through this conditioning, at inference time DISYRE assumes in practice that test images are highly anomalous, and iteratively uses the modelled score function to generate a restoration. Such an assumption is not realistic and hampers its ability to identify subtle anomalies, as confirmed by the reported results, where this approach is outperformed by previous methods. Moreover, if the model is instead run assuming lower levels of anomaly severity (acting on the value used for $t$), the overall performance drops significantly.

\section{Methods}

\label{sections:method}

\subsection{Disentangled Anomaly Generation}
We hypothesize that the sensitivity of DISYRE to the assumed severity is due to the choice of FPI \cite{FPI} as the strategy for generating synthetic anomalies. Since FPI anomalies are created by interpolating two image patches, this is unlikely to produce significant shifts in intensities at the beginning of the corruption process. This feature of FPI can lead to models associating low $t$ only to texture anomalies and, consequently, negatively impacting the restoration abilities of the final model for hyper- and hypo-intense lesions when conditioning with a low $t$.

To address this issue, we introduce a more expressive Disentangled Anomaly Generation (DAG) process, which disentangles 3 attributes of synthetic anomalies: shape, texture, and intensity bias, where intensity bias is defined as a shift in the expected tissue intensities. When generating anomalies each of the attributes is independently randomly sampled following a uniform distribution and combined to generate a synthetic anomaly. Differently from FPI, DAG can create gradual hyper- and hypo-intense synthetic lesions, which we believe can make the cold-diffusion pipeline robust to the assumed severity during inference. Additionally, by generating anomalies with different combinations of anomalous textures and intensities, DAG can increase the coverage of the anomalous image distribution.

\textbf{Anomaly Shape:} We follow the implementation from~\cite{MOOD22} where the shape component is determined by a mask $m \in \mathbb{R}^{H \times W}$, with $H$ and $W$ respectively height and width of the image, and $m_i \in [0,1]$, where pixels with $m_i = 0$ are uncorrupted and $m_i = 1$ are anomalous. Masks are generated randomly and their edges smoothed so anomalies blend gradually with surrounding areas.

\textbf{Anomaly Texture: } The texture component is created following FPI~\cite{FPI}. Image patches randomly sampled from the healthy training set, referred to as \textit{foreign patches} $x_{fp} \in \mathbb{R}^{H \times W}$, are used to replace sections of healthy training images $x \in \mathbb{R}^{H \times W}$. The corruption is achieved with a convex combination of $x$ and $x_{fp}$ governed by the interpolation factor $\alpha_{text} \in [0,1]$. Differently from FPI, we attempt to separate bias and texture components by normalizing $x_{fp}$ so it has the same range of intensities as the original image $x$ in the anomaly location (being $\overline{x}_{fp}$ normalized version of $x_{fp}$). Texture corrupted images $x_{text}$ are defined as: 
\begin{equation}
x_{text} = (1 - \alpha_{text} \cdot m) \cdot x + \alpha_{text} \cdot m \cdot \overline{x}_{fp}
\label{eq:xt_t}
\end{equation}

\textbf{Intensity Bias:} Intensity corruptions are achieved shifting intensities in the location determined by $m$ by a factor $\pm \alpha_{bias}$. To obtain more plausible anomalies we aim to restrict the shift of intensities to specific tissue types (e.g. white or gray matter in the case of Brain MRI) within the anomaly location determined by $m$. We propose to use k-Means to identify clusters of intensities across the whole dataset which we assume to represent tissue types ($K=5$ clusters are used in our experiments). When generating anomalies, we randomly choose the tissue type $k$ and define a tissue type mask as $m_{tissue} \in \mathbb{R}^{H \times W}$ with elements $m_{tissue}^i = 1$ if $x^i$ is assigned to cluster $k$ and $m_{tissue}^i = 0$ otherwise. Additionally, the randomly sampled ${bias}_{sign} \in \{-1,1\}$  defines whether intensities will be decreased or increased. The bias corruption is applied to $x_{text}$ to obtain the final synthetically corrupted image $x_{sc}$  used during training: 
\begin{equation}
x_{sc} = (1 -  m \cdot m_{tissue}) \cdot x_{text} + (x_{text} + \alpha_{bias} \cdot {bias}_{sign}) \cdot m \cdot m_{tissue}
\label{eq:xt_b}
\end{equation}

Figure \ref{fig:anom_gen} shows anomalies generated when varying attributes $\alpha_{bias}$ and $\alpha_{text}$.

\subsection{Cold-diffusion with synthetic anomaly corruptions}
In diffusion models~\cite{DDPM}, a corruption schedule $B(t)$ defines a gradual Gaussian noise corruption process which starts at time step $t=0$, where images are unaltered, and ends at $t=T$, where only Gaussian noise remains. Diffusion models are trained to reverse this process: by denoising samples while conditioned on $t$, they learn to generate high-frequency details for low values of $t$ and structural features for high values. Cold-diffusion \cite{cold_diffusion} follows the same strategy but proposes alternative corruptions not based on noise.

We implement a synthetic anomaly cold-diffusion method to generate restorations from images with anomalies following DISYRE~\cite{DISYRE}. In the original implementation, the noise schedule $B(t)$ from DDPMs is adopted and used to set the FPI~\cite{FPI} interpolation factor $\alpha = B(t)$. Consequently, when $t=0$, $\alpha_t = 0$ the image belongs to the observed healthy distribution, while when $t=T$, $\alpha_t = 1$ and the image is highly anomalous. A network $P_\theta$ is trained to restore corrupted images $x_t$ into their original counterparts $x_0$ when conditioning on $t$, $x_0 \approx \hat{x}_0 = P_{\theta}(x_t,t)$. Network parameters are therefore optimized using the objective:
\begin{equation}
\mathbb{E}_{t \sim [1,T],x_0,x_{fp}} (\| x_0 - P_{\theta}(x_t,t) \|^2)
\label{eq:loss}
\end{equation}

We propose to extend the model to the 2 dimensions of corruption in DAG, i.e. $\alpha_{text}$ and $\alpha_{bias}$, by simply conditioning on $t = B^{-1}\left(max(\alpha_{text},\alpha_{bias})\right)$, where $B^{-1}$ is the inverse of $B(t)$, so the model receives a single indication of severity. 

\subsection{Anomaly localization}
Diffusion models generate images iteratively, starting at step $t=T$ with Gaussian noise $x_T \sim \mathcal{N}(0, I)$, where  $x_T \in \mathbb{R}^{H \times W}$. At each of the $T$ iterations the diffusion model \textit{denoises} the sample until obtaining an image from the observed distribution at step $t=0$. Similarly, \cite{DISYRE} proposes to iteratively use the score function learned in the domain of synthetically corrupted images to \textit{heal} test images $x$ and obtain restorations. The residuals between restorations and original images are then used as anomaly score (AS). Importantly, by starting the restoration process with $t=T$ the method assumes a high severity of abnormality. We refer to the AS proposed in \cite{DISYRE} as \textit{multi-step} in our experiments. 

In this work, we propose an effective alternative, which is to ensemble single-step restorations conditioned at different degrees of abnormality. 
\begin{equation}
AS_{ensemble} = \sum^0_{t=T}(\|P_{\theta}(x,t) - x\|)
\label{eq:as_e}
\end{equation}

Our proposed Ensemble AS allows us to leverage the increased model robustness from DAG. 
Similarly to \cite{DISYRE}, in practice we skip steps based on a \textit{step size} hyper-parameter. We choose $step\_size = 25$ as a compromise between coverage of the anomalous distribution and inference speed, and report results in the supplementary materials. As in \cite{DISYRE}, we use sliding window inference and weight the AS predicted for individual patches using the proportion of foreground. Sliding window inference allows to apply our pipeline seamlessly to other high-dimensional modalities. In our experiments we use a patch-overlap of 0.25 and \textit{gaussian weighting} of patches~\cite{nnUNet}.

\section{Experiments}
\label{sections:results}
\newcommand{\dice}{$\lceil$Dice$\rceil$}
\subsection{Experimental setup}
Our experiments use the Brain MRI setup from~\cite{UPD} which includes datasets:
\begin{itemize}
    \item Cambridge Centre for Ageing and Neuroscience (CamCAN) \cite{CamCAN}: 653 T1- and T2-weighted images of healthy and lesion free subjects. CamCAN dataset is used for training ($N=603$) and to evaluate training convergence ($N=50$).
    \item Anatomical Tracings of Lesions After Stroke (ATLAS) dataset \cite{ATLAS}: 655 T1-weighted images of stroke patients with  annotated lesions. Some of these images are affected by artefacts and inadequate skull-stripping, which UAD methods often identify as anomalies despite not being annotated as such and thus lead to performance underestimation. For our internal ablations we use the subset excluding issues (ATLAS-ex, N=571), but report the results for the full dataset when comparing to previous studies~\cite{UPD}.
    \item Multimodal Brain Tumor Segmentation (BraTS) Challenge dataset \cite{brats} 2020 edition: 369 T1- and T2-weighted images from subjects with gliomas, provided with manual annotations of lesions. 

\end{itemize}
\begin{figure*}[b!]
    \centering
    \includegraphics[width=1\columnwidth]{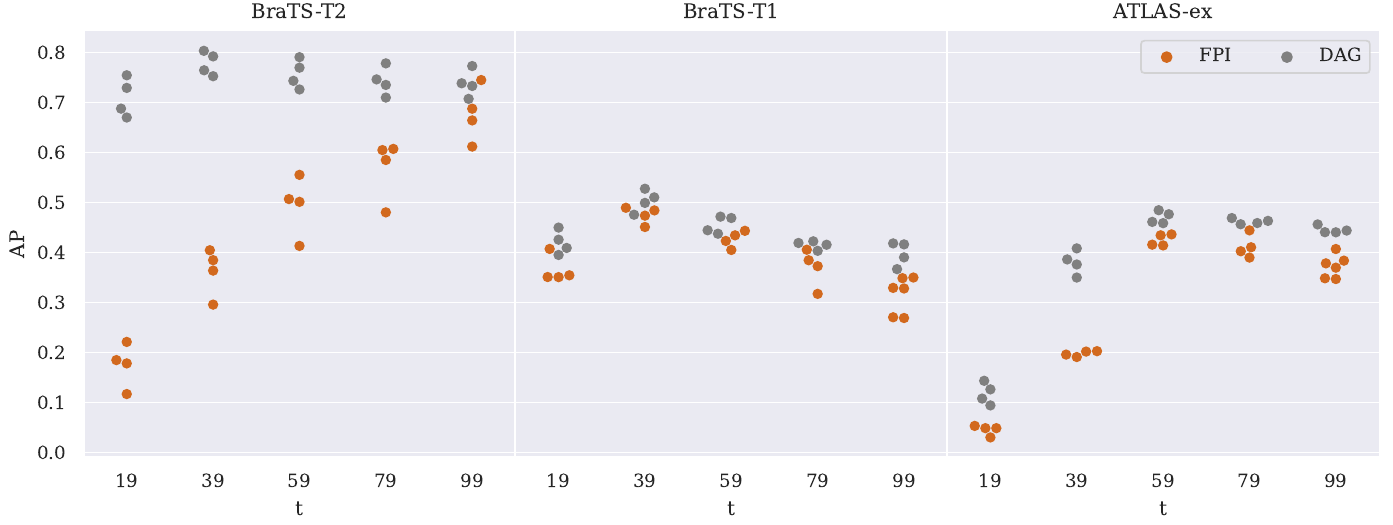}
    \caption{AP for 4 independent seeds, using single-step restoration, FPI vs DAG.}
    \label{fig:profiles}
\end{figure*}

ATLAS and BraTS datasets are used to evaluate UAD performance while CamCAN is used as the training, normative dataset. 
We follow the pre-processing specified in~\cite{UPD}. Similarly to \cite{DISYRE}, we used a 2D patch-based pipeline with patches of $128 \times 128$ size obtained from non-empty axial slices. Our network implementation, diffusion corruption schedule, and training schedule follow the specifications in \cite{DISYRE} but we ran training for longer (200,000 steps) to ensure convergence.

\subsection{Results and Discussion} 
To quantitatively evaluate our method, we use Average Precision (AP) and an estimate of the best possible S\o rensen-Dice index (\dice) as in~\cite{UPD}. 
To evaluate the robustness of the model to the assumed severity ($t$), we obtained AS from single-step restorations, i.e.  $\| x - P(x,t) \|$ for different $t$.  Figure \ref{fig:profiles} shows that DAG improves results across all values of $t$. The improvement is particularly relevant when conditioning with $t=19$ and $t=39$ (i.e. low severity assumed) in BraTS-T2 and ATLAS datasets. Since anomalies in BraTS-T2 and ATLAS appear hyper- and hypo-intense respectively, these results showcase that, by enabling the creation of low-grade bias-only anomalies, DAG addresses the limitations of FPI and makes the model robust across
different severity levels.
\newcommand{\api}{AP\textsubscript{i}}
\newcommand{\app}{AP}
\newcommand{\row}[5]{}  
\begin{table}[t!]
    \centering
    \caption{Localization results of the image-reconstruction (\textbf{IR}), feature-modeling (\textbf{FM}), self-supervised (\textbf{S-S}) and hybrid (\textbf{HY}) methods. Best scores are bold.}
    \resizebox{0.65\columnwidth}{!}{
    \begin{tabular}{clcccccc}
        & \multicolumn{1}{c}{} & \multicolumn{2}{c}{\textbf{ATLAS}} & \multicolumn{2}{c}{\textbf{BraTS-T1}} & \multicolumn{2}{c}{\textbf{BraTS-T2}} \\
        \cmidrule(lr){3-4} \cmidrule(lr){5-6} \cmidrule(lr){7-8}
        & Method & \app & \dice & \app & \dice & \app & \dice \\
        \midrule
        \multirow{2}{*}{\rotatebox[origin=c]{90}{\textbf{IR}}}
        & VAE \cite{Baur2020} & 0.11 & 0.20 & 0.13 & 0.19 & 0.28 & 0.33 \\
        & r-VAE \cite{Chen2020} & 0.09 & 0.17 & 0.13 & 0.19 & 0.36 & 0.40 \\
        \midrule
        \multirow{4}{*}{\rotatebox[origin=c]{90}{\textbf{FM}}}
        & FAE \cite{FAE} & 0.08 & 0.18 & 0.42 & 0.45 & 0.51 & 0.52 \\
        & PaDiM \cite{padim}& 0.05 & 0.13 & 0.21 & 0.28 & 0.34 & 0.38 \\
        & CFLOW-AD \cite{CFFLOW}& 0.04 & 0.10 & 0.16 & 0.24 & 0.31 & 0.35 \\
        & RD \cite{deng2022anomaly} & 0.11 & 0.22 & 0.36 & 0.42 & 0.47 & 0.50 \\
        \midrule
        \multirow{4}{*}{\rotatebox[origin=c]{90}{\textbf{S-S}}}
        & PII~\cite{PII} & 0.03 & 0.07 & 0.13 & 0.22 & 0.13 & 0.22 \\
        & DAE~\cite{DAE} & 0.05 & 0.13 & 0.13 & 0.20 & 0.47 & 0.49 \\
        & CutPaste \cite{li2021cutpaste} & 0.03 & 0.06 & 0.07 & 0.13 & 0.22 & 0.26 \\
        & MOOD22 \cite{MOOD22} & 0.10 & 0.21 & 0.24 & 0.31 & 0.47 & 0.48 \\
        \midrule
        \multirow{2}{*}{\rotatebox[origin=c]{90}{\textbf{HY}}}
        & DISYRE~\cite{DISYRE} & 0.29 & 0.37 & 0.34 & 0.41 & 0.75 & 0.70 \\
        & DISYRE v2 (Ours) & \textbf{0.33} & \textbf{0.45} & \textbf{0.48} & \textbf{0.51} & \textbf{0.79} & \textbf{0.73} \\
        \bottomrule
    \end{tabular}}
    \label{tab:main}
\end{table}

Next we evaluated the interaction between proposed AG and AS strategies in Table~\ref{tab:ablation}. To further evaluate the impact of conditioning, we also include models trained to generate restorations without conditioning on $t$ altogether, with their AS defined as  $\| x - P(x) \|$. Conditional setups show overall better performance compared to unconditional ones, showcasing the benefits of the diffusion-like pipeline. However, prior methods (i.e. multi-step AS) fail to improve the unconditional setting in BraTS-T1 (where anomalies are often subtle) by assuming a high severity during inference time. DISYRE v2 solves this limitation with the proposed Ensemble AS, which raises the AP by 10\% (0.48 vs 0.44). The interaction between the proposed AS-AG strategies is noteworthy: DAG makes single-step restorations robust to the assumed severity, benefiting the Ensembled AS when combining predictions.

\begin{table}[t]\setlength{\tabcolsep}{8pt}
    \centering
    \caption{AP in test sets for combinations of Anomaly Score (AS) and Anomaly Generation (AG). Mean and std. dev. on 4 seeds.}
    \label{tab:ablation}
    \resizebox{0.9\columnwidth}{!}{
    \begin{tabular}{ccccccc}
   & AS & AG & \textbf{ATLAS-ex}& \textbf{BraTS-T1}&\textbf{BraTS-T2}\\
         \midrule
         & Unconditional & FPI & 0.36 ± 0.03 &   0.44 ± 0.01 & 0.68 ± 0.00 \\
         & Unconditional & DAG & 0.40 ± 0.01 &   0.44 ± 0.01 & 0.74 ± 0.01 \\
         \midrule
         & Multi-step & FPI & 0.42 ± 0.02 &  0.40 ± 0.02 & 0.73 ± 0.04 \\
         & Multi-step & DAG & 0.47 ± 0.00 &  0.44 ± 0.02 & 0.78 ± 0.02 \\
         & Ensemble & FPI & 0.44 ± 0.02&   0.46 ± 0.02 & 0.69 ± 0.06  \\
         & Ensemble & DAG & \textbf{0.47± 0.01} & \textbf{0.48 ± 0.01} & \textbf{0.79 ± 0.02} \\
        \bottomrule
    \end{tabular}
    }
\end{table}

\begin{figure*}
    \centering
    \includegraphics[width=1.\columnwidth]{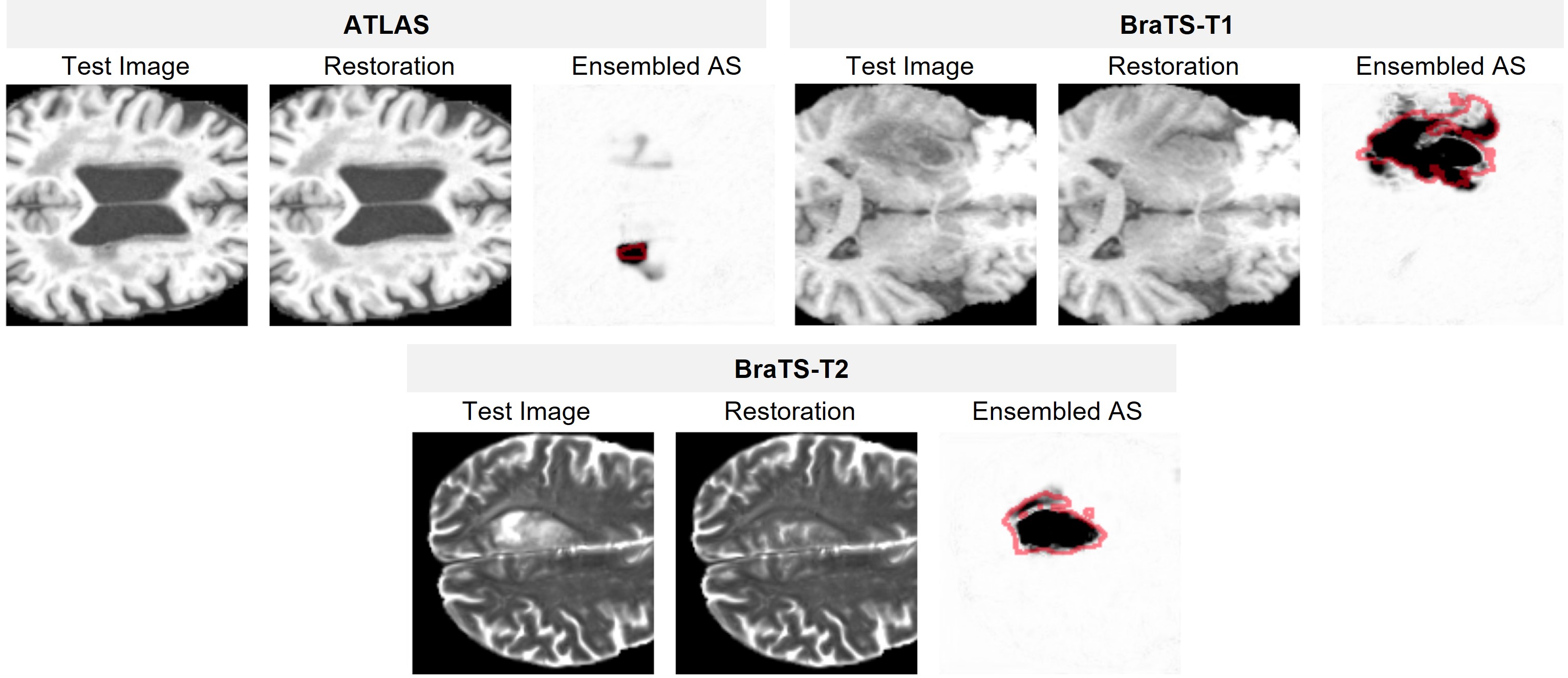}
    \caption{Examples of restorations and Ensembled AS  vs ground truth (red outline).}
    \label{fig:qualitatives}
\end{figure*}

Finally, Table \ref{tab:main} includes a comparison of our method with baselines evaluated in~\cite{UPD} and DISYRE~\cite{DISYRE}. The performance of hybrid methods showcases that the combination of image restoration and anomaly localization tasks improve above their individual components, \emph{i.e.} image-reconstruction and synthetic-anomaly localization \cite{PII,li2021cutpaste}. By addressing the described limitations of the original DISYRE, DISYRE v2 outperforms it by a good margin across the 3 tasks. 
Figure \ref{fig:qualitatives} includes qualitative examples of predictions. Additional comparisons between Ensemble and Multi-step AS can be found in the supplementary materials.

\section{Conclusion}
In this work we presented DISYRE v2, a novel evolution of DISYRE, and evaluated its performance in a common test bench of Brain MRI lesions. We set the new state-of-the-art in the three tasks where we evaluated it, and importantly, improved the performance metrics for the two more challenging tasks (ATLAS and BraTS-T1) by more than 10\%.
In our analysis we identify limitations of previous works and present as a solution a novel combination of synthetic AG, AS and cold-diffusion modelling which allows to make fewer assumptions at inference time, removing biases towards specific anomalies. Consequently, DISYRE v2 not only improves performance in the tasks evaluated, but it is expected to be more robust across the diverse and challenging naturally occurring medical anomalies, setting a new state-of-the-art in medical UAD.

\noindent\textbf{Acknowledgments: } B. Kainz received funding from the ERC project MIA-NORMAL 101083647; M. Baugh is supported by a UKRI DTP award.

\subsubsection{\discintname}
The authors have no competing interests to declare that are relevant to the content of this article.
%
%
%
\bibliographystyle{splncs04}
\bibliography{micc24}

\end{document}